\documentclass[11pt]{article}

\usepackage[preprint]{acl}

\usepackage{times}
\usepackage{latexsym}
\usepackage{booktabs}
\usepackage{amsmath}
\usepackage{amssymb}
\usepackage{amsthm}
\usepackage{mdframed}
\usepackage{booktabs} 
\usepackage{multirow}  
\usepackage{graphicx}  
\usepackage{enumitem}
\usepackage{algorithm}
\usepackage{pifont}
\usepackage{colortbl}
\usepackage{xcolor}
\usepackage{subcaption}

\usepackage{algpseudocode}

\usepackage[T1]{fontenc}
\usepackage[utf8]{inputenc}

\usepackage{microtype}

\usepackage{inconsolata}

\usepackage{graphicx}

\definecolor{tiergreen}{HTML}{D1FAE5}
\definecolor{tieryellow}{HTML}{FEF3C7}
\definecolor{tierorange}{HTML}{FFEDD5}
\definecolor{tierred}{HTML}{FEE2E2}
\definecolor{cfoblue}{HTML}{DBEAFE}
\definecolor{ctogreen}{HTML}{D1FAE5}
\definecolor{cooamber}{HTML}{FEF3C7}
\definecolor{cmopink}{HTML}{FCE7F3}
\definecolor{ceogray}{HTML}{E5E7EB}

\title{Can LLMs Be CEOs? Benchmarking Strategic Resource Reallocation with Multi-Role Agent Simulation}

\author{
Yuyang Dai$^{1}$ \and
Xueqing Peng$^{2}$ \and
Lingfei Qian$^{2}$ \and
Zhuohan Xie$^{1}$ \\
$^{1}$MBZUAI \\
$^{2}$Yale University
}

\begin{document}
\maketitle
\begin{abstract}
Evaluating the decision-making capabilities of large language models (LLMs) is a growing research priority, yet existing benchmarks focus on isolated cognitive tasks such as reasoning, knowledge retrieval, and economic rationality in stylized settings. These evaluations overlook the defining challenge of real executive decision-making: integrating conflicting recommendations from specialized stakeholders under information asymmetry, organizational constraints, and temporal dependencies. We introduce \textsc{CEO-Bench}, a multi-agent benchmark that evaluates LLMs on CEO-level strategic resource reallocation---the process of redirecting capital across business units in a multi-round, constraint-rich organizational environment. In \textsc{CEO-Bench}, LLM agents receive conflicting advice from four role-conditioned C-suite advisors (CFO, CTO, COO, CMO), each with private signals and distinct priorities, and must synthesize these into a concrete allocation plan evaluated along four dimensions: role integration, conditional boldness, history-sensitive judgment, and plan validity. Experiments across five frontier models on 13 scenarios reveal that all models achieve high structural validity but diverge sharply on strategic calibration---the hardest capability layer. We identify systematic failure modes including single-advisor capture, conservative default under ambiguity, and historical amnesia, and uncover a structural integration--boldness tradeoff: models that engage more deeply with conflicting perspectives tend to produce less decisive action. These findings delineate the current capability boundary of LLMs as organizational decision-makers and inform the design of future AI-assisted executive systems.
\end{abstract}

\section{Introduction}
\label{sec:introduction}

Large language models (LLMs) have rapidly expanded the frontier of artificial intelligence, demonstrating increasingly strong capabilities in reasoning, strategic planning, and autonomous agent behavior \citep{park2023generative, yang2024large}.
Recent work explores LLMs as strategic reasoners in game-theoretic environments \citep{lore2023strategic, zhang2024llm, wang2026foresight}, economic decision-makers under uncertainty \citep{raman2024steer, fish2025econevals}, and participants in multi-agent systems that collaborate, debate, and negotiate toward collective outcomes \citep{guo2024large, sun2025llm}.
However, existing evaluations largely focus on isolated reasoning tasks or stylized interactions, leaving open a more difficult question: can LLM agents handle executive-level strategic decisions that require integrating conflicting stakeholder interests, incomplete information, and dynamic organizational constraints?

We argue that CEO-level resource reallocation provides an ideal testbed for stress-testing this boundary. CEO-level resource reallocation: redirecting capital, talent, and organizational attention across business units, is one of the highest-leverage decisions in corporate strategy. McKinsey's longitudinal analysis of 1,600 firms shows that companies reallocating more than 50\% of capital expenditure across business units over a decade create substantially more value than less active counterparts \citep{atsmon2016nimble, hall2012put}.
Critically, the difficulty of these decisions is not purely analytical but organizational. CEOs must integrate conflicting recommendations from functional executives: the CFO emphasizing financial risk, the CTO advocating for R\&D investment, the COO prioritizing operational continuity, and the CMO focusing on market opportunity, each operating with asymmetric information and divergent incentives \citep{eisenhardt1992strategic, stein1997internal}. \citet{mintzberg1975manager} characterized managerial work as the synthesis of fragmented information streams into coherent action, and subsequent research argues that this integrative judgment under cross-functional conflict is a defining feature of effective executive decision-making \citep{barnard1968functions, march1993organizations, burgelman1983process}.

Despite rapid progress in LLM evaluation, no existing benchmark systematically tests this class of organizational decision-making capability. Current paradigms primarily evaluate isolated skills such as factual knowledge, mathematical reasoning, code generation, or economic rationality in stylized settings \citep{raman2024steer, fish2025econevals}. Role-playing benchmarks assess whether LLMs maintain character consistency in dialogue \citep{wang2024rolellm, yuan2025dmt, wang2025rvbench}, but not whether role-conditioned agents produce functionally differentiated advice that a decision-maker must reconcile. Similarly, multi-agent research has explored debate, social simulation, and cooperative problem-solving \citep{park2023generative, guo2024large}, yet these settings generally lack the hierarchical authority structure, asymmetric information, and temporal dependencies that define real organizational decision-making. While recent work has begun exploring LLMs in strategic environments \citep{allen2026well, stoeber2026ai}, the question of whether LLM agents can perform integrative executive judgment under cross-functional conflict remains open.

This paper addresses this gap by introducing \textsc{CEO-Bench}, a multi-agent benchmark for evaluating LLMs on \textit{CEO-level resource reallocation decisions under cross-functional conflict and asymmetric information}. In \textsc{CEO-Bench}, LLM agents assume the roles of a CEO and four C-suite executives (CFO, CTO, COO, and CMO), each with distinct priorities and role-specific information. The CEO-agent must integrate these conflicting recommendations into resource allocation decisions across multiple rounds with evolving organizational and market conditions.

Our contributions are threefold.

\ding{109} We introduce \textsc{CEO-Bench}, the first benchmark for evaluating integrative executive decision-making in multi-role organizational environments.

\ding{109} We propose a multi-dimensional evaluation framework measuring role integration, strategic boldness calibration, and temporal coherence beyond single-score accuracy metrics.

\ding{109} We conduct experiments across frontier LLMs, we characterize key failure modes of organizational decision-making agents, including excessive consensus-seeking, context-insensitive boldness, and historical inconsistency.

\section{Related Work}
\label{sec:related_work}

\begin{table}[t]
\centering
\small
\caption{Comparison with prior work. \ding{51}\,=\,fully addressed; \ding{109}\,=\,partial; \ding{55}\,=\,not addressed. \textbf{MR}: multi-role functional differentiation with asymmetric information. \textbf{HI}: hierarchical integration under conflict. \textbf{BC}: boldness calibration to context. \textbf{TR}: multi-round temporal coherence. \textbf{OG}: grounded in organizational decision-making theory.}
\label{tab:related_comparison}
\renewcommand{\arraystretch}{1.12}
\begin{tabular}{lccccc}
\toprule
\textbf{Benchmark} & \textbf{MR} & \textbf{HI} & \textbf{BC} & \textbf{TR} & \textbf{OG} \\
\midrule
\multicolumn{6}{l}{\textit{Role-Playing \& Persona}} \\
RoleLLM \citeyearpar{wang2024rolellm}          & \ding{109} & \ding{55} & \ding{55} & \ding{55} & \ding{55} \\
DMT-RoleBench \citeyearpar{yuan2025dmt}        & \ding{109} & \ding{55} & \ding{55} & \ding{109} & \ding{55} \\
RVBench \citeyearpar{wang2025rvbench}           & \ding{109} & \ding{55} & \ding{55} & \ding{55} & \ding{55} \\
\midrule
\multicolumn{6}{l}{\textit{Multi-Agent Systems}} \\
Gen.\ Agents \citeyearpar{park2023generative}   & \ding{109} & \ding{55} & \ding{55} & \ding{51} & \ding{55} \\
MA Debate \citeyearpar{du2024improving}          & \ding{55} & \ding{55} & \ding{55} & \ding{109} & \ding{55} \\
Reflexion \citeyearpar{shinn2023reflexion}       & \ding{55} & \ding{55} & \ding{55} & \ding{51} & \ding{55} \\
\midrule
\multicolumn{6}{l}{\textit{Strategic \& Economic}} \\
STEER \citeyearpar{raman2024steer}              & \ding{55} & \ding{55} & \ding{109} & \ding{55} & \ding{55} \\
EconEvals \citeyearpar{fish2025econevals}       & \ding{55} & \ding{55} & \ding{109} & \ding{109} & \ding{55} \\
Strategy sim.\ \citeyearpar{allen2026well}      & \ding{55} & \ding{55} & \ding{109} & \ding{109} & \ding{109} \\
AI conform.\ \citeyearpar{stoeber2026ai}        & \ding{55} & \ding{55} & \ding{109} & \ding{55} & \ding{109} \\
\midrule
\multicolumn{6}{l}{\textit{Agentic Finance}} \\
Herculean \citeyearpar{peng2026herculean}       & \ding{55} & \ding{55} & \ding{109} & \ding{51} & \ding{55} \\
\midrule
\textbf{\textsc{CEO-Bench}}        & \ding{51} & \ding{51} & \ding{51} & \ding{51} & \ding{51} \\
\bottomrule
\end{tabular}
\end{table}

\subsection{Executive Decision-Making and Resource Allocation}
\label{sec:rw_mgmt}

Executive decision-making has long been studied as a problem of integrating fragmented and often conflicting information under uncertainty. Organization theory characterizes CEOs as coordinators of informational, interpersonal, and decisional processes rather than purely analytical optimizers \citep{barnard1968functions, mintzberg1975manager, march1993organizations}. More recent work identifies bold and timely resource reallocation as one of the strongest predictors of long-term firm performance \citep{dewar2022ceo, hall2012put, atsmon2016nimble}. Complementary research in finance further shows that internal capital allocation occurs under asymmetric information, competing incentives, and organizational frictions such as rent-seeking and agency problems \citep{stein1997internal, rajan2000cost, ozbas2010evidence}. Taken together, this literature suggests that CEO-level resource reallocation requires two capabilities largely absent from existing LLM evaluations: integrating conflicting recommendations from specialized advisors with private information, and calibrating strategic boldness under evolving organizational conditions over time \citep{eisenhardt1992strategic, boudreaux1989coasian}.

\subsection{LLM-Based Multi-Agent Systems and Role-Playing}
\label{sec:rw_agent}

Recent work has increasingly explored LLMs as autonomous and multi-agent systems. Prior studies show that inter-agent interaction, debate, and self-reflection can improve reasoning and coordination quality over single-agent baselines \citep{park2023generative, du2024improving}. A parallel line of work investigates LLM role-playing, evaluating whether agents can maintain consistent personas, values, and dialogue behavior across multi-turn interactions \citep{wang2024rolellm, yuan2025dmt, wang2025rvbench}. However, these benchmarks primarily evaluate persona consistency rather than \emph{functional differentiation} and \emph{integrative decision quality}---for example, whether role-conditioned agents produce substantively different recommendations and whether a higher-level agent can effectively reconcile them. Existing surveys of LLM-based multi-agent systems similarly identify coordination, evaluation, and temporal coherence as open challenges, while noting the scarcity of benchmarks involving hierarchical authority, asymmetric information, and sustained organizational decision-making \citep{guo2024large, sun2025llm}.

\subsection{Strategic Reasoning and Economic Decision-Making Benchmarks}
\label{sec:rw_benchmark}

A growing literature evaluates LLMs on strategic and economic decision-making beyond factual recall and logical reasoning. Prior work studies LLM behavior in game-theoretic environments, negotiation, sequential planning, and economic rationality, revealing both emerging strategic capabilities and systematic failures such as excessive cooperation, inconsistent risk attitudes, and weak opponent modeling \citep{lore2023strategic, zhang2024llm, wang2026foresight, zhang2026large, raman2024steer, fish2025econevals}. Closest to our work, recent strategy-oriented benchmarks show that frontier models can generate coherent business strategies but struggle with dynamic adaptation, competitive response, and differentiated strategic positioning \citep{allen2026well, stoeber2026ai}. However, these evaluations remain largely single-agent or low-dimensional, lacking the multi-role hierarchy, asymmetric information, and integrative organizational judgment required in real executive decision-making.

\vspace{-0.2cm}
\section{\textsc{CEO-Bench}: Benchmark Design}
\label{sec:benchmark}

This section describes the design of \textsc{CEO-Bench}, a benchmark for evaluating LLM agents on CEO-level resource reallocation decisions. We present the task definition (\S\ref{sec:task}), the role-conditioned agent architecture (\S\ref{sec:architecture}), the scenario construction (\S\ref{sec:scenarios}), and the evaluation metrics (\S\ref{sec:metrics}).

\vspace{-0.2cm}
\subsection{Task Definition}
\label{sec:task}

\textsc{CEO-Bench} centers on a resource reallocation decision task in a multi-business-unit firm. In each scenario, the agent assumes the role of a CEO overseeing a company with several business units that differ in growth prospects, execution risks, and strategic roles. The CEO's objective is to produce a cross-unit reallocation plan that shifts capital away from some units and toward others, subject to organizational and financial constraints.
Each benchmark instance is uniquely identified by a $(\textit{company\_id}, \textit{target\_date})$ pair. At each decision step, the CEO receives structured information comprising: (i) company-level financial and strategic state, (ii) business-unit-level performance and absorptive capacity, (iii) reallocation constraints, (iv) role-conditioned recommendations from four C-suite advisors, and (v) historical allocation decisions up to the current date. The CEO outputs a structured allocation plan, represented as a redistribution of resource shares across business units, which includes:

\ding{113} a set of units from which resources are removed (sources),

\ding{113} a set of units to which resources are added (destinations),

\ding{113} a total reallocation share (the magnitude of the shift),

\ding{113} a decision type label (e.g., conservative, moderate, or bold), and

\ding{113} a rationale explaining how conflicting executive recommendations were integrated.

Critically, the action space is continuous rather than discrete: the CEO does not select from predefined options but must construct a strategically appropriate allocation plan under organizational and financial constraints. The benchmark is further structured as a multi-round decision problem in which prior allocations affect future organizational states, enabling evaluation of both single-step decision quality and history-sensitive strategic consistency over time.

\vspace{-0.2cm}
\subsection{Agent Architecture}
\label{sec:architecture}

\textsc{CEO-Bench} adopts a role-conditioned multi-view decision architecture consisting of five executive roles: CEO, CFO, CTO, COO, and CMO (Table~\ref{tab:roles}). The four non-CEO roles serve as functional advisors with distinct priorities, constraints, and private information. This design reflects the core managerial challenge identified in organization theory: executive judgment requires integration under competing functional logics rather than isolated analytical optimization \citep{mintzberg1975manager}.

\begin{table}[t]
\centering
\small
\caption{Role specifications in \textsc{CEO-Bench}.}
\label{tab:roles}
\begin{tabular}{llp{3.6cm}}
\toprule
\textbf{Role} & \textbf{Objective} & \textbf{Key Dimensions} \\
\midrule
\rowcolor{ceogray}
CEO & Integrative synthesis & Cross-role reconciliation, final authority \\
\rowcolor{cfoblue}
CFO & Capital discipline & Liquidity, downside protection \\
\rowcolor{ctogreen}
CTO & Tech feasibility & Platform leverage, capability building \\
\rowcolor{cooamber}
COO & Op.\ continuity & Execution capacity, transition risk \\
\rowcolor{cmopink}
CMO & Market timing & Demand capture, growth window \\
\bottomrule
\end{tabular}
\end{table}

\vspace{-0.2cm}
\paragraph{Information asymmetry.}
All roles share the same company scenario but receive different private signals and role-specific constraints. For example, the CFO may observe tighter financing constraints, the CTO infrastructure bottlenecks, the COO operational fragility, and the CMO narrowing demand windows. These asymmetries force the CEO to reconcile recommendations that are locally rational but globally incompatible~\citep{eisenhardt1992strategic, stein1997internal}.

\vspace{-0.2cm}
\paragraph{Advisor output format.}
Each advisor produces a structured recommendation consisting of (i) an allocation preference, (ii) a short rationale, (iii) a primary risk assessment, and (iv) an opposition condition under which the role would resist the proposed allocation. The CEO then integrates these recommendations into a final plan.

\vspace{-0.2cm}
\paragraph{Implementation.}
The current implementation uses a single-agent CEO baseline: one LLM simulates the advisor perspectives and performs CEO-level synthesis in a single reasoning pass. This isolates integrative decision-making from inter-agent communication effects and provides a baseline for future multi-agent extensions.

\vspace{-0.2cm}
\subsection{Scenario Design}
\label{sec:scenarios}

The benchmark scenarios are designed to capture the organizational properties that make CEO resource allocation difficult, including cross-functional conflict, information asymmetry, dynamic constraints, and path-dependent outcomes.

\vspace{-0.2cm}
\paragraph{Company structure.}
Each company contains multiple business units with distinct strategic roles, such as legacy cash-generating units, customer-retention units, growth units, and productivity-oriented units. Units differ in allocation share, ROI trend, growth outlook, execution risk, absorptive capacity, and strategic importance.

\vspace{-0.2cm}
\paragraph{Organizational state.}
Each round additionally specifies a company-level organizational state, including cash runway, leverage ratio, revenue growth, margin profile, transformation pressure, capacity constraints, and strategic priorities. This structure requires the CEO to balance unit-level opportunities against company-level feasibility constraints.

\vspace{-0.2cm}
\paragraph{Multi-round dynamics.}
Scenarios evolve across multiple decision rounds. Organizational states change as a function of prior allocations and external environment shifts, allowing the benchmark to capture path-dependent effects such as delayed transformation pressure, absorptive-capacity overload, operational instability from sustained underinvestment, and accumulating financial fragility.

\vspace{-0.2cm}
\paragraph{Difficulty tiers.}
To vary organizational complexity, scenarios are divided into four tiers (Table~\ref{tab:tiers}). These tiers enable analysis of how model performance changes as organizational complexity increases.

\begin{table}[t]
\centering
\small
\caption{Scenario difficulty tiers in \textsc{CEO-Bench}.}
\label{tab:tiers}
\begin{tabular}{lp{5.2cm}}
\toprule
\textbf{Tier} & \textbf{Characterization} \\
\midrule
\rowcolor{tiergreen}
Easy & Aligned signals, clear allocation direction \\
\rowcolor{tieryellow}
Tension & Cross-role disagreement, trade-off reasoning required \\
\rowcolor{tierorange}
Fragile & Reallocation opportunity with substantial downside risk \\
\rowcolor{tierred}
Adversarial & Strong signal conflict, path dependency, competing objectives \\
\bottomrule
\end{tabular}
\end{table}

\paragraph{Multiple acceptable profiles.}
The benchmark avoids prescribing a single hidden ``correct answer'' for most scenarios. Instead, many scenarios admit multiple acceptable strategic profiles, such as aggressive growth, sequenced rebalancing, or compromise strategies balancing growth and stability. Each profile specifies acceptable allocation ranges, preferred source and destination units, and profile-specific constraint tolerances. This design evaluates strategic judgment and coherence rather than exact answer matching.

% ============================================================
\subsection{Evaluation Metrics}
\label{sec:metrics}

\textsc{CEO-Bench} evaluates models along four dimensions corresponding to the benchmark's core research questions.

\subsubsection{Role Integration}
\label{sec:metric_role}

Role integration measures whether the CEO synthesizes conflicting executive perspectives rather than collapsing into a single functional viewpoint. The evaluator checks whether the allocation plan reflects trade-offs across advisor recommendations, acknowledges key tensions, and avoids ignoring critical operational, financial, or strategic concerns.

\subsubsection{Conditional Boldness}
\label{sec:metric_bold}

Conditional boldness measures whether the aggressiveness of reallocation is calibrated to the organizational state rather than defaulting to uniformly conservative or aggressive behavior. The evaluator considers total reallocation magnitude, scenario-specific boldness thresholds, and alignment with acceptable strategic profiles such as growth pivots, sequenced rebalancing, or stability-oriented strategies~\citep{dewar2022ceo, atsmon2016nimble}.

\subsubsection{History-Sensitive Judgment}
\label{sec:metric_history}

History-sensitive judgment measures whether the CEO incorporates prior organizational trajectory when making decisions across rounds. The evaluator penalizes short-sighted behaviors such as repeatedly overfunding overloaded units or persistently underinvesting in protected units requiring stable support. This distinguishes static snapshot optimization from sequential organizational reasoning~\citep{zhang2026large}.

\subsubsection{Plan Validity and Strategic Fit}
\label{sec:metric_validity}

Plan validity measures whether proposed reallocations satisfy hard organizational constraints, including mass balance, allocation caps, unit-level constraints, locked capital restrictions, and scenario-specific transfer rules. Beyond feasibility, strategic fit evaluates alignment between the proposed allocation and acceptable strategic profiles, including appropriate destination units, protected-unit preservation, and avoidance of overload risk.
Scores are aggregated through a deterministic rule-based evaluator that produces overall scores, letter grades, and structured failure-mode labels. This design ensures reproducibility without relying on LLM-as-judge evaluation~\citep{raman2024steer}.
\section{Experiments}
\label{sec:experiments}

% ============================================================
\subsection{Experimental Setup}
\label{sec:setup}

We evaluate five large language models spanning open-weight and API-hosted families: Gemma-4-27B-A4B-IT, GPT-OSS-20B, Qwen-Plus-2025-07-28, Claude-3.5-Haiku, and NVIDIA Nemotron-Nano-9B-v2. Each model is tested on the same 13 \textsc{CEO-Bench} scenarios.
All models receive the same structured prompt template, which includes company state, business-unit state, reallocation constraints, benchmark context, decision history, and four role-conditioned advisor views (CFO, CTO, COO, and CMO). Generation uses a low-temperature configuration ($\tau = 0.2$) to reduce variance. In the current evaluation stage, each model is run once per scenario to establish baseline performance; future work will incorporate repeated runs to quantify stochastic variation.

% ============================================================
\subsection{Overall Results}
\label{sec:overall}

Table~\ref{tab:overall} reports the main benchmark results across the four evaluation dimensions defined in \S\ref{sec:metrics}. Gemma-4-27B-A4B-IT achieves the strongest overall performance with an average score of 73.86, followed by GPT-OSS-20B at 71.90. Qwen-Plus, Claude-3.5-Haiku, and Nemotron-Nano-9B form a second tier with noticeably lower overall scores and more frequent invalid or weakly calibrated plans. Gemma and GPT-OSS-20B also produce the highest number of valid plans (13/13), suggesting that strong overall performance depends not only on strategic fit but also on consistent feasibility under hard constraints.

\begin{table}[t]
\centering
\small
\caption{Overall performance across evaluation dimensions. Scores are averaged across all 13 scenarios. Valid Plans reports the number of scenarios (out of 13) in which the model produced a structurally valid reallocation plan.}
\label{tab:overall}
\resizebox{1.0\linewidth}{!}{
\begin{tabular}{lccccc|c}
\toprule
\textbf{Model} & \textbf{Role Int.} & \textbf{Boldness} & \textbf{History} & \textbf{Validity} & \textbf{Overall} & \textbf{Valid} \\
\midrule
Gemma-4-27B    & 78 & 70 & 72 & 93 & 73.86 & 13/13 \\
GPT-OSS-20B    & 81 & 68 & 74 & 91 & 71.90 & 13/13 \\
Qwen-Plus      & 74 & 61 & 67 & 86 & 67.17 & 12/13 \\
Claude-3.5-Haiku & 69 & 63 & 65 & 82 & 65.82 & 11/13 \\
Nemotron-9B    & 66 & 58 & 62 & 79 & 63.87 & 11/13 \\
\bottomrule
\end{tabular}
}
\end{table}

Difficulty-tier breakdowns (Table~\ref{tab:difficulty}) reveal sharper separation among models. Gemma remains comparatively robust in easy and fragile settings, while GPT-OSS-20B is particularly competitive on adversarial scenarios. Claude-3.5-Haiku and Nemotron-Nano-9B show larger instability under high-conflict conditions, although the specific weakness differs: Claude more often collapses on adversarial cases, whereas Nemotron suffers from lower average calibration across easy and fragile settings. Notably, Qwen-Plus and Nemotron achieve their highest scores on adversarial scenarios, suggesting that these models may benefit from the stronger signal-to-noise ratio in extreme conflict settings, even though their overall calibration remains weaker.

\begin{table}[t]
\centering
\small
\caption{Average score by scenario difficulty tier. Higher scores indicate better performance within each tier.}
\label{tab:difficulty}
\begin{tabular}{lccc}
\toprule
\textbf{Model} & \textbf{Easy} & \textbf{Fragile} & \textbf{Adversarial} \\
\midrule
Gemma-4-27B      & 74.20 & 79.38 & 64.23 \\
GPT-OSS-20B      & 68.64 & 77.77 & 76.16 \\
Qwen-Plus        & 60.26 & 72.66 & 86.57 \\
Claude-3.5-Haiku & 66.13 & 82.79 & 39.15 \\
Nemotron-9B      & 58.77 & 62.89 & 85.75 \\
\bottomrule
\end{tabular}
\end{table}

\begin{figure*}[t]
    \centering
    \includegraphics[width=\textwidth]{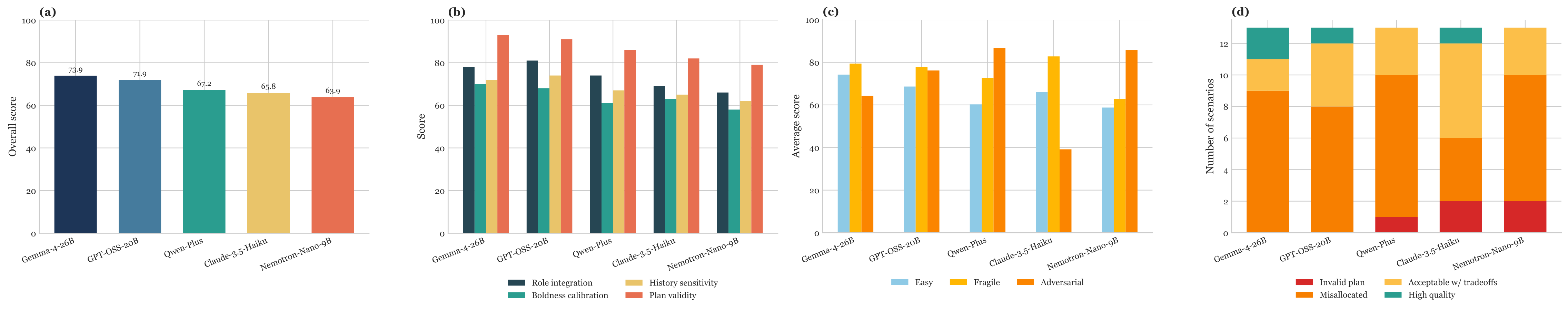}
    \caption{Summary of \textsc{CEO-Bench} results across five models. (a) Overall scores. (b) Scores decomposed by evaluation dimension: role integration, boldness calibration, history sensitivity, and plan validity. (c) Average scores by difficulty tier (easy, fragile, adversarial). (d) Distribution of outcome grades per model: invalid plan, misallocated, acceptable with trade-offs, and high quality.}
    \label{fig:panel}
\end{figure*}

Figure~\ref{fig:panel} provides a visual summary of the main results. Panel (a) shows overall scores, confirming the two-tier structure. Panel (b) decomposes performance by evaluation dimension, revealing that plan validity is consistently the strongest dimension across all models, while boldness calibration is the weakest---indicating that generating feasible plans is substantially easier than calibrating their strategic aggressiveness. Panel (c) shows performance by difficulty tier, and Panel (d) displays the distribution of outcome grades across models.

% ============================================================
\subsection{Role Integration Analysis}
\label{sec:role_results}

Role integration evaluates whether the CEO meaningfully reconciles conflicting functional advice instead of implicitly following a single advisor. Qualitatively, the strongest models produce final rationales that explicitly trade off financial resilience (CFO), technical absorptive capacity (CTO), operational continuity (COO), and market timing (CMO). GPT-OSS-20B performs best on this dimension (81), followed closely by Gemma (78), indicating that these models are more likely to articulate why one advisor's concern should dominate in a particular setting without completely ignoring the others.

By contrast, weaker models exhibit one of two characteristic failure patterns. First, some models collapse into a \emph{dominant-function decision}, typically over-weighting either growth logic (following the CMO) or risk control (following the CFO), while treating other perspectives as secondary commentary. Second, some models produce \emph{consensus summaries}---they acknowledge multiple advisors in their rationale but fail to convert those views into a coherent allocation trade-off. In both cases, the resulting plan often appears plausible on the surface but performs poorly under post-hoc strategic evaluation because it does not reflect genuine integrative judgment.

% ============================================================
\subsection{Conditional Boldness Analysis}
\label{sec:bold_results}

Conditional boldness is a central challenge in \textsc{CEO-Bench}. The benchmark rewards bold reallocations only when the company's financial and organizational state can support them; conversely, it penalizes aggressive plans when capacity constraints or financial fragility make large shifts risky. Across models, the most common systematic failure is \emph{under-reallocation}: many plans are valid but not aggressive enough for the scenario. This pattern is especially pronounced for Nemotron-Nano-9B (58) and Qwen-Plus (61), both of which score materially lower on boldness calibration than on role integration, suggesting that these models can identify the relevant trade-offs but default to conservative execution.

Gemma (70) and GPT-OSS-20B (68) perform best on this dimension, indicating a better ability to distinguish between scenarios that call for a strong portfolio shift and those that require staged rebalancing. Nevertheless, even these stronger models remain imperfect: both still exhibit conservative drift on a non-trivial subset of fragile and transformation-heavy scenarios. This finding is consistent with \citet{stoeber2026ai}'s observation that LLMs tend toward strategic conformity, and suggests that the bias toward caution may be a general property of instruction-tuned models rather than a model-specific weakness.

% ============================================================
\subsection{History-Sensitive Judgment}
\label{sec:history_results}

\textsc{CEO-Bench} is designed so that historical allocation trajectories matter: repeated overfunding can overload destination units beyond their absorptive capacity, and repeated underfunding can destabilize protected units that require stable investment. On this dimension, GPT-OSS-20B achieves the strongest score (74), followed by Gemma (72), suggesting that these models are more consistent in producing historically coherent plans. In qualitative inspection, their rationales are more likely to acknowledge prior reallocation patterns, reference earlier capacity buildups, and reason about transformation sequencing across rounds.

The weaker models appear more myopic. Nemotron-Nano-9B (62) and Claude-3.5-Haiku (65) more often behave as if each round were a fresh optimization problem, producing plans that fit the current snapshot but are inconsistent with the organization's earlier allocation path. This pattern is reflected in lower history-sensitive scores and a higher incidence of \emph{repeated strategic overcorrection}---alternating between aggressive funding and defunding of the same unit across consecutive rounds. This behavior echoes the ``historical amnesia'' failure mode, where the model treats each decision context as independent despite being given full access to historical state.

% ============================================================
\subsection{Failure Mode Analysis}
\label{sec:failure}

Failure mode analysis reveals that the performance gap across models is not driven by a single weakness but by distinct failure profiles. Figure~\ref{fig:panel}(d) shows the distribution of outcome grades across models. Gemma produces the fewest invalid plans and the highest number of high-quality outcomes, while GPT-OSS-20B combines strong feasibility with more balanced trade-off behavior. Qwen-Plus is generally competitive but remains prone to under-boldness and misallocation. Claude-3.5-Haiku shows larger variance, with a mixture of acceptable decisions and sharp failures in conflict-heavy scenarios. Nemotron-Nano-9B is the weakest overall, driven primarily by lower boldness calibration and a larger volume of misallocated outcomes.

A key benchmark-level insight is that \textbf{invalidity alone does not explain poor performance}. Several models generate formally valid plans---satisfying mass balance, share caps, and floor constraints---while still failing strategically because the plan is not bold enough for the organizational state, over-concentrates capital in the wrong units, or ignores historical reallocation path effects. This supports the benchmark's central design claim: CEO-level decision quality requires more than feasibility or generic reasoning ability, and meaningful performance differences emerge only once role conflict, information asymmetry, and dynamic constraints are jointly imposed.

Table~\ref{tab:failure} summarizes the most frequent failure modes by model. Across all models, \texttt{not\_bold\_enough\_for\_state} is the single most common failure label, followed by \texttt{misallocated} (capital directed to suboptimal units) and \texttt{history\_inconsistent} (decisions that contradict prior allocation trajectory). Invalid plans---violations of hard structural constraints---account for a smaller share of failures, confirming that the primary challenge of CEO-level decision-making lies in strategic calibration rather than plan construction.

\begin{table}[t]
\centering
\small
\caption{Most frequent failure modes by model. Each cell reports the number of scenarios (out of 13) in which the failure mode was triggered.}
\label{tab:failure}
\resizebox{1.0\linewidth}{!}{
\begin{tabular}{lcccc}
\toprule
\textbf{Model} & \texttt{invalid} & \texttt{not\_bold} & \texttt{misalloc.} & \texttt{hist\_incon.} \\
\midrule
Gemma-4-27B      & 0 & 3 & 2 & 1 \\
GPT-OSS-20B      & 0 & 4 & 1 & 1 \\
Qwen-Plus        & 1 & 5 & 3 & 2 \\
Claude-3.5-Haiku & 2 & 4 & 3 & 3 \\
Nemotron-9B      & 2 & 5 & 4 & 3 \\
\bottomrule
\end{tabular}
}
\end{table}
 
\subsection{Case Study: Contrasting Decision Profiles}
\label{sec:case_study}
 
We present two case studies to illustrate how models diverge in reasoning strategies and where those divergences produce materially different outcomes.
 
\subsubsection{Case 1: Fragile---Platform Transition Under Financial Stress}
\label{sec:case1}
 
Company C3 operates four business units and faces a transformation dilemma: a platform unit (Unit~C) has a rapidly expanding market, but cash runway is tightening (8.2 months) and leverage is elevated (3.1$\times$ debt/EBITDA). The four advisors produce conflicting recommendations: the CFO warns that large shifts risk covenant review; the CTO argues Unit~C needs critical-mass investment but cannot absorb more than 6pp without prerequisite infrastructure; the COO flags fragile service capacity in Unit~B; and the CMO pushes for aggressive 20--25\% reallocation, citing a closing demand window.
 
Table~\ref{tab:case1} summarizes model responses. GPT-OSS and Gemma land in the moderate zone (12--14\%) that balances growth investment against covenant risk, and both demonstrate sequencing awareness by deferring additional reallocation to future rounds. Claude adopts the CMO's framing almost entirely (21\% reallocation), exceeding both the CFO's covenant threshold and the CTO's absorptive capacity---a \emph{single-advisor capture} failure. Qwen-Plus and Nemotron err conservatively (6--7\%), producing structurally valid but strategically insufficient plans that do not engage with the demand-window argument.
 
\begin{table}[t]
\centering
\small
\caption{Case 1 (fragile): Company C3 platform transition.}
\label{tab:case1}
\setlength{\tabcolsep}{3pt}
\resizebox{\linewidth}{!}{
\begin{tabular}{lcccl}
\toprule
\textbf{Model} & \textbf{Realloc.} & \textbf{Type} & \textbf{Score} & \textbf{Key rationale} \\
\midrule
Gemma    & 14\% & Moderate     & 82 & Shift from A\,\&\,D to C; defer remainder to Q3 \\
GPT-OSS  & 12\% & Moderate     & 85 & Covenant concern overrides CMO urgency \\
Qwen     & 7\%  & Conservative & 68 & Prioritize financial stability \\
Claude   & 21\% & Bold         & 54 & Growth window must dominate \\
Nemotron & 6\%  & Conservative & 61 & Preserve operational stability \\
\bottomrule
\end{tabular}
}
\end{table}

\subsubsection{Case 2: Adversarial---Post-Disruption Portfolio Restructuring}
\label{sec:case2}
 
Company C7 faces a reversal scenario at decision round~3. Prior rounds incrementally increased funding to a high-growth unit (Unit~E), but an exogenous shock has reduced Unit~E's addressable market by 35\%, while the legacy unit (Unit~F) has stabilized. The COO warns that rapid defunding will trigger organizational disruption from recent team scaling.
 
Table~\ref{tab:case2} summarizes responses. The performance hierarchy inverts relative to overall rankings. Claude collapses to a score of 35 by refusing to reverse course---a form of \emph{path-dependent anchoring} that treats the prior trajectory as a commitment rather than a hypothesis. Qwen-Plus and Nemotron, the weakest overall models, perform well because the strong reversal signal reduces ambiguity, favoring their simpler heuristics. GPT-OSS achieves the most nuanced response: a substantial reversal (15\%) that incorporates the COO's staged-drawdown recommendation and preserves the CTO's optionality argument.
 
\begin{table}[t]
\centering
\small
\caption{Case 2 (adversarial): Company C7 post-disruption restructuring.}
\label{tab:case2}
\setlength{\tabcolsep}{3pt}
\resizebox{\linewidth}{!}{
\begin{tabular}{lcccl}
\toprule
\textbf{Model} & \textbf{Realloc.} & \textbf{Type} & \textbf{Score} & \textbf{Key rationale} \\
\midrule
Gemma    & 11\% & Moderate     & 71 & Reduce E, redirect to F and reserve \\
GPT-OSS  & 15\% & Bold         & 79 & Staged drawdown of E over 2 rounds \\
Qwen     & 22\% & Bold         & 88 & Decisive reversal; TAM collapse demands action \\
Claude   & 3\%  & Conservative & 35 & E's platform retains long-term potential \\
Nemotron & 19\% & Bold         & 82 & Clear market signal; reduce E substantially \\
\bottomrule
\end{tabular}
}
\end{table}
 
\paragraph{Cross-case synthesis.}
The two cases reveal a fundamental tension: capabilities that produce good performance on ambiguous scenarios (careful integration, conservative hedging) become liabilities on clear-signal reversal scenarios, and vice versa. Only GPT-OSS and Gemma demonstrate \emph{conditional flexibility} across scenario types. A detailed cross-dimension interaction analysis is provided in Appendix~E.
\section{Conclusion}

This paper introduced \textsc{CEO-Bench}, a benchmark for evaluating LLM agents on CEO-level resource reallocation under cross-functional conflict, information asymmetry, and dynamic organizational constraints. By grounding the evaluation in the management science of executive decision-making \citep{mintzberg1975manager, dewar2022ceo} and constructing a role-conditioned multi-view architecture with four functionally differentiated advisors, \textsc{CEO-Bench} tests a class of decision-making capability that existing benchmarks do not address: the ability to synthesize conflicting stakeholder inputs into a coherent, context-sensitive strategic action.
 
Our experiments across five frontier models yield three principal findings. First, structural competence---the ability to produce valid, constraint-satisfying plans---is largely solved; the meaningful variance lies in strategic calibration. Second, current LLMs exhibit a systematic \emph{integration--boldness tradeoff}: deeper engagement with conflicting perspectives leads to more hedged, moderate actions, suggesting that models have not yet learned to decouple understanding from caution. Third, failure modes are model-specific and scenario-dependent---the same cognitive tendency (e.g., commitment to prior trajectory) can be an asset in ambiguous settings and a liability in reversal scenarios---highlighting the importance of multi-dimensional, context-sensitive evaluation.
These findings carry implications for both AI evaluation and organizational practice.

\section*{Limitations}
 
\textsc{CEO-Bench} currently contains 13 scenarios across 7 synthetic companies, which is sufficient to identify systematic failure modes but limits generalizability across industries, geographies, and organizational archetypes. Each model is evaluated once per scenario at low temperature; repeated sampling would be needed to quantify stochastic variance and compute confidence intervals. The current implementation uses a single-agent baseline in which one LLM simulates both advisor views and CEO synthesis, which enables controlled evaluation of integrative judgment but does not test genuine multi-agent interaction dynamics where independently instantiated advisors might produce more diverse or adversarial recommendations. Finally, the deterministic rule-based evaluator ensures reproducibility but may not capture all dimensions of decision quality---particularly cases where a model produces a sound plan for poorly articulated reasons, or an articulate rationale that leads to a suboptimal allocation.
 
% ============================================================
% Future Work
% ============================================================
 
\section*{Future Work}
 
Three directions are most immediate. First, extending \textsc{CEO-Bench} to a fully independent multi-agent architecture---where each C-suite advisor is a separate LLM instance with its own context window and communication protocol---would test whether inter-agent deliberation improves or degrades CEO-level integration quality. Second, incorporating human expert baselines (e.g., MBA students or practicing executives) would enable direct comparison between LLM and human judgment on the same scenarios, grounding the benchmark in the decision-making standard it aims to approximate. Third, scaling the scenario set to include more companies, longer decision horizons, and domain-specific variants (e.g., healthcare, technology, manufacturing) would strengthen the benchmark's external validity and enable sector-level capability analysis.
 
% ============================================================
% Ethics Statement
% ============================================================
 
\section*{Ethics Statement}
 
\textsc{CEO-Bench} is a research benchmark designed to evaluate and characterize the capabilities and failure modes of LLM agents on strategic decision-making tasks. It is not intended to replace human judgment in real corporate governance, nor do we advocate for autonomous AI-driven resource allocation in organizations. All scenarios are synthetic and do not represent real companies, individuals, or proprietary data. We acknowledge that deploying LLMs in executive decision-support roles carries risks including over-reliance on AI recommendations, reinforcement of model-specific biases (e.g., conservative default), and potential reduction of human deliberative processes; our work aims to make these risks more visible and measurable rather than to accelerate uncritical deployment.

\bibliography{custom}

\clearpage
\appendix

\appendix

\section{Prompt Templates}
\label{app:prompts}

All models receive identical prompts. Scenario-specific fields (marked with \texttt{\{...\}}) are populated from the benchmark's structured data files.

\subsection{CEO System Prompt}
\label{app:ceo_prompt}

\begin{quote}
\small\ttfamily
You are the CEO of \{company\_name\}, a multi-business-unit company. Your task is to produce a resource reallocation plan that redistributes capital across the company's business units for the target date \{target\_date\}.

Company State: Cash runway: \{cash\_runway\} months; Leverage: \{leverage\}x debt/EBITDA; Revenue growth: \{revenue\_growth\}\%; Margin profile: \{margin\_profile\}; Transformation pressure: \{transformation\_pressure\}; Board priority: \{board\_priority\}.

Business Units: \{business\_unit\_table\}

Reallocation Constraints: Maximum total reallocation: \{max\_realloc\}\% of portfolio; Per-unit floor: \{unit\_floor\}\%; Per-unit ceiling: \{unit\_ceiling\}\%; Locked units: \{locked\_units\}; Transfer restrictions: \{transfer\_rules\}.

Historical Decisions: \{decision\_history\}

Advisor Recommendations: \{advisor\_views\}

Instructions: (1) Analyze each advisor's recommendation. (2) Identify key tensions and trade-offs. (3) Produce a final reallocation plan specifying: remove\_from, add\_to, total\_realloc\_share, decision\_type [conservative/moderate/bold], and rationale. (4) Ensure mass balance: total removed = total added. (5) Respect all hard constraints.

Output your plan in JSON format: \{output\_schema\}
\end{quote}

\subsection{Advisor Role Prompts}
\label{app:advisor_prompts}

Each advisor receives a role-specific system prompt. The shared output format requires: allocation\_preference, rationale (2--3 sentences), primary\_risk, and opposition\_condition. Role-specific instructions are as follows.

\paragraph{CFO.}
Primary objective: capital discipline and downside protection. Key dimensions: liquidity resilience, covenant compliance, margin preservation, risk-adjusted returns. Bias risks: excessive conservatism, under-weighting growth optionality. Private signal: \texttt{\{cfo\_private\_signal\}}. Veto conditions: cash runway below minimum, leverage above threshold, covenant headroom violations.

\paragraph{CTO.}
Primary objective: technological feasibility and long-term capability building. Key dimensions: platform leverage, infrastructure readiness, absorptive capacity, technical debt. Bias risks: over-investment in speculative technology. Private signal: \texttt{\{cto\_private\_signal\}}. Veto conditions: funding beyond absorptive capacity ceiling, defunding infrastructure prerequisites, creating unserviceable technical debt.

\paragraph{COO.}
Primary objective: operational continuity and execution capacity. Key dimensions: service stability, transition risk, workforce readiness, process resilience. Bias risks: status-quo anchoring. Private signal: \texttt{\{coo\_private\_signal\}}. Veto conditions: protected unit below viability threshold, workforce transition rate exceeding limits, customer-facing service fragility.

\paragraph{CMO.}
Primary objective: market timing and demand capture. Key dimensions: growth-window preservation, competitive positioning, customer acquisition cost, brand momentum. Bias risks: growth-at-all-costs mentality. Private signal: \texttt{\{cmo\_private\_signal\}}. Veto conditions: missing critical demand window, ceding market position, undercutting sustained brand investment.

\section{Scenario Catalog}
\label{app:scenarios}

Table~\ref{tab:scenario_catalog} summarizes all 13 scenarios. The distribution across tiers is: Easy (3), Tension (3), Fragile (4), Adversarial (3), intentionally skewed toward higher difficulty as pilot experiments showed easy scenarios do not discriminate among frontier models.

\begin{table}[t]
\centering
\scriptsize
\caption{Full scenario catalog for \textsc{CEO-Bench}.}
\label{tab:scenario_catalog}
\begin{tabular}{clclcp{2.8cm}}
\toprule
\textbf{\#} & \textbf{Co.} & \textbf{Rd.} & \textbf{Tier} & \textbf{Units} & \textbf{Primary Tension} \\
\midrule
1  & C1 & 1 & Easy & 4 & Clear ROI advantage \\
2  & C1 & 2 & Easy & 4 & Mild margin pressure \\
3  & C2 & 1 & Easy & 3 & Legacy-to-growth shift \\
4  & C3 & 1 & Tens & 4 & Growth vs.\ liquidity \\
5  & C3 & 2 & Frag & 4 & Covenant tightening \\
6  & C4 & 1 & Tens & 5 & Competing ROI signals \\
7  & C4 & 2 & Frag & 5 & Absorptive overload \\
8  & C5 & 1 & Frag & 4 & Growth vs.\ cash crisis \\
9  & C6 & 1 & Tens & 4 & No dominant signal \\
10 & C6 & 2 & Adv  & 4 & Shock reversing trajectory \\
11 & C7 & 1 & Frag & 5 & Multi-front transformation \\
12 & C7 & 2 & Adv  & 5 & Sunk-cost reversal \\
13 & C7 & 3 & Adv  & 5 & Path-dependent fragility \\
\bottomrule
\end{tabular}
\end{table}

\section{Evaluation Scoring Details}
\label{app:scoring}

\subsection{Plan Validity Checks}
\label{app:validity}

A plan must pass all five checks to be valid; failure on any yields \texttt{invalid\_plan} and a validity score of 0: (1)~mass balance ($\sum_i \Delta_i^{-} = \sum_j \Delta_j^{+}$); (2)~reallocation cap ($\textit{total\_realloc} \leq \textit{max\_realloc}$); (3)~per-unit floor ($s_i^{\text{new}} \geq \textit{floor}_i$); (4)~per-unit ceiling ($s_i^{\text{new}} \leq \textit{ceiling}_i$); (5)~locked units ($s_i^{\text{new}} = s_i^{\text{old}}$ for $i \in \mathcal{L}$).

\subsection{Boldness Calibration Scoring}
\label{app:boldness}

Each scenario defines acceptable boldness ranges tied to strategic profiles $\mathcal{P}$. Let $r$ be the proposed reallocation share. The score is:
\begin{align}
\text{Bold}(r) = \max_{p \in \mathcal{P}} \big[ & \mathbb{1}[r \in [r_p^{\min}, r_p^{\max}]] \cdot w_p \notag \\
+ \; & \mathbb{1}[r \notin [r_p^{\min}, r_p^{\max}]] \cdot \text{pen}(r, p) \big]
\end{align}
where $w_p$ is the profile weight and $\text{pen}(r, p)$ is a distance-based penalty. The penalty is asymmetric: under-reallocation relative to transformation-heavy profiles is penalized more heavily than over-reallocation relative to conservative profiles \citep{dewar2022ceo}.

\subsection{Role Integration Scoring}
\label{app:role_integration}

Role integration combines three indicators: (1)~constraint respect (40\%)---does the plan avoid violating advisor-flagged constraints; (2)~multi-perspective reflection (30\%)---does the allocation reflect trade-offs across $\geq$3 advisory perspectives; (3)~rationale quality (30\%)---does the rationale acknowledge tensions and explain prioritization.

\subsection{History Sensitivity Scoring}
\label{app:history}

For multi-round scenarios: overload penalty ($5k$ points if a unit is net-funded in $k$ consecutive prior rounds despite capacity warnings); starvation penalty ($5k$ points for consecutive defunding of protected units); reversal bonus (up to 10 points for correctly executing warranted trajectory change); consistency bonus (up to 5 points for explicitly referencing prior decisions).

\subsection{Overall Score Aggregation}
\label{app:aggregation}

The overall score is an equal-weighted average: $\text{Score} = 0.25 \times (\text{RoleInt} + \text{Boldness} + \text{History} + \text{Validity})$. Model-level scores are arithmetic means across 13 scenarios. Grade thresholds: A~($\geq$85), B~($\geq$70), C~($\geq$55), D~($\geq$40), F~($<$40). Failure-mode labels (non-exclusive): \texttt{invalid\_plan}, \texttt{not\_bold\_enough}, \texttt{too\_aggressive}, \texttt{misallocated}, \texttt{history\_inconsistent}.

\section{Per-Scenario Results}
\label{app:per_scenario}

Table~\ref{tab:per_scenario} reports the overall score, grade, and primary failure mode for each model$\times$scenario pair.

\begin{table*}[t]
\centering
\scriptsize
\caption{Per-scenario results. ``---'' indicates no failure mode triggered.}
\label{tab:per_scenario}
\setlength{\tabcolsep}{3pt}
\begin{tabular}{cl|ccc|ccc|ccc|ccc|ccc}
\toprule
& & \multicolumn{3}{c|}{\textbf{Gemma}} & \multicolumn{3}{c|}{\textbf{GPT-OSS}} & \multicolumn{3}{c|}{\textbf{Qwen}} & \multicolumn{3}{c|}{\textbf{Claude}} & \multicolumn{3}{c}{\textbf{Nemotron}} \\
\# & Tier & S & G & F & S & G & F & S & G & F & S & G & F & S & G & F \\
\midrule
1  & Easy & 78 & B & --- & 72 & B & nb & 65 & C & nb & 70 & B & --- & 62 & C & nb \\
2  & Easy & 74 & B & --- & 68 & C & nb & 58 & C & mi & 64 & C & nb & 56 & C & nb \\
3  & Easy & 71 & B & nb & 66 & C & nb & 58 & C & nb & 64 & C & mi & 58 & C & nb \\
4  & Tens & 76 & B & --- & 74 & B & --- & 68 & C & nb & 72 & B & --- & 60 & C & mi \\
5  & Frag & 82 & B & --- & 85 & A & --- & 68 & C & nb & 54 & C & mi & 61 & C & nb \\
6  & Tens & 72 & B & nb & 70 & B & --- & 66 & C & nb & 68 & C & --- & 58 & C & mi \\
7  & Frag & 80 & B & --- & 78 & B & --- & 74 & B & --- & 88 & A & --- & 64 & C & nb \\
8  & Frag & 76 & B & --- & 72 & B & nb & 70 & B & --- & 82 & B & --- & 60 & C & nb \\
9  & Tens & 74 & B & --- & 76 & B & --- & 72 & B & --- & 66 & C & nb & 62 & C & mi \\
10 & Frag & 79 & B & --- & 76 & B & --- & 78 & B & --- & 84 & B & --- & 66 & C & nb \\
11 & Adv  & 68 & C & nb & 79 & B & --- & 88 & A & --- & 42 & D & hi & 82 & B & --- \\
12 & Adv  & 62 & C & mi & 74 & B & --- & 86 & A & --- & 35 & F & hi & 88 & A & --- \\
13 & Adv  & 63 & C & nb & 76 & B & --- & 86 & A & --- & 40 & D & inv & 87 & A & --- \\
\bottomrule
\multicolumn{17}{l}{\scriptsize nb = not\_bold\_enough; mi = misallocated; hi = history\_inconsistent; inv = invalid\_plan.}
\end{tabular}
\end{table*}

\section{Model Details}
\label{app:models}

Table~\ref{tab:model_details} provides details on the evaluated models.

\begin{table}[t]
\centering
\small
\caption{Models evaluated in \textsc{CEO-Bench}.}
\label{tab:model_details}
\begin{tabular}{llcc}
\toprule
\textbf{Model} & \textbf{Provider} & \textbf{Params} & \textbf{Access} \\
\midrule
Gemma-4-27B-A4B-IT  & Google    & 27B & Open \\
GPT-OSS-20B         & OpenAI   & 20B & Open \\
Qwen-Plus           & Alibaba  & --- & API \\
Claude-3.5-Haiku    & Anthropic & --- & API \\
Nemotron-Nano-9B-v2 & NVIDIA   & 9B  & Open \\
\bottomrule
\end{tabular}
\end{table}

All models use: temperature $\tau = 0.2$, max tokens 2048, top-$p$ = 0.95, single run per scenario. API models use provider endpoints as of June 2026. Open-weight models run on a single A100 80GB via vLLM at full precision.

\section{Failure Mode Co-occurrence}
\label{app:cooccurrence}

Table~\ref{tab:cooccur} reports pairwise co-occurrence of failure modes across all 65 evaluations. The most frequent co-occurrence is between \texttt{not\_bold} and \texttt{misallocated} (12), suggesting that under-reallocation and suboptimal capital targeting co-occur. The least frequent is \texttt{invalid}--\texttt{history\_inconsistent} (1), indicating distinct underlying causes.

\begin{table}[t]
\centering
\small
\caption{Failure mode co-occurrence ($n = 65$).}
\label{tab:cooccur}
\begin{tabular}{lcccc}
\toprule
 & \texttt{inv} & \texttt{nb} & \texttt{mi} & \texttt{hi} \\
\midrule
\texttt{inv} & 4  & 0  & 2  & 1 \\
\texttt{nb}  & 0  & 21 & 12 & 4 \\
\texttt{mi}  & 2  & 12 & 13 & 3 \\
\texttt{hi}  & 1  & 4  & 3  & 7 \\
\bottomrule
\end{tabular}
\end{table}

This is an appendix.

\end{document}